\documentclass[acmsmall]{acmart}

\usepackage{subfig}
\usepackage{amsmath}
\usepackage{mathtools}
\usepackage{algorithmic}
\usepackage{graphicx}
\usepackage{transparent}
\usepackage{xcolor}
\usepackage{listings}
\usepackage{url}
\usepackage{multirow}

\usepackage{fancyhdr}
\usepackage{tikz}
\usetikzlibrary{tikzmark}

\usetikzlibrary{arrows.meta, positioning, quotes}

\usepackage{pdfpages}
\def\BibTeX{{\rm B\kern-.05em{\sc i\kern-.025em b}\kern-.08emT\kern-.1667em\lower.7ex\hbox{E}\kern-.125emX}}

\begin{document}

\setcopyright{none}
\acmJournal{JETC}
\acmPrice{}

\title{A Resource-Efficient Embedded Iris Recognition System Using Fully Convolutional Networks}

%
\author{Hokchhay Tann}
\email{hokchhay_tann@alumni.brown.edu}
\orcid{0003-4762-4972}
\author{Heng Zhao}
\email{heng_zhao@alumni.brown.edu}
\author{Sherief Reda}
\email{sherief_reda@brown.edu}
\affiliation{%
  \institution{Brown University}
  \streetaddress{182 Hope Street}
  \city{Providence}
  \state{Rhode Island}
  \postcode{02912}
}

%
\begin{abstract}
Applications of Fully Convolutional Networks (FCN) in iris segmentation have shown promising advances. For mobile and embedded systems, a significant challenge is that the proposed FCN architectures are extremely computationally demanding. In this article, we propose a resource-efficient, end-to-end iris recognition flow, which consists of FCN-based segmentation, contour fitting, followed by Daugman normalization and encoding. To attain accurate and efficient FCN models, we propose a three-step SW/HW co-design methodology consisting of FCN architectural exploration, precision quantization, and hardware acceleration. In our exploration, we propose multiple FCN models, and in comparison to previous works, our best-performing model requires 50$\times$ less FLOPs per inference while achieving a new state-of-the-art segmentation accuracy. Next, we select the most efficient set of models and further reduce their computational complexity through weights and activations quantization using 8-bit dynamic fixed-point (DFP) format. Each model is then incorporated into an end-to-end flow for true recognition performance evaluation. A few of our end-to-end pipelines outperform the previous state-of-the-art on two datasets evaluated. Finally, we propose a novel DFP accelerator and fully demonstrate the SW/HW co-design realization of our flow on an embedded FPGA platform. In comparison with the embedded CPU, our hardware acceleration achieves up to 8.3$\times$ speedup for the overall pipeline while using less than 15\% of the available FPGA resources. We also provide comparisons between the FPGA system and an embedded GPU showing different benefits and drawbacks for the two platforms.
\end{abstract}

%
%

 

\maketitle

\section{Introduction}
Due to the unique and rich signatures in the irises of each individual, iris recognition has been shown as one of the most secure forms of biometric identification \cite{daugman1993}. Unlike other biometric features such as fingerprints and voice, the irises hardly change over the course of an individual's lifetime. Recently, iris recognition becomes increasingly common on various wearable and mobile devices. For these systems, high level of security and efficient recognition processing pipeline with low computational complexity are the two stringent requirements for deployment.

A variety of algorithms and implementations have been proposed over the years for iris recognition pipelines \cite{wildes1994, poursaberi2005, xu2008, daugman2009iris}. For typical processing flows, some of the main difficulties include obtaining quality iris image and accurately segmenting the iris region. For iris segmentation, several algorithms have been developed \cite{caht, wahet, gst, masek, osiris, irisseg} using a diverse set of techniques such as circular Hough transform and integrodifferential operator. With the recent success of deep learning, emerging studies on iris recognition adopt various forms of Deep Neural Networks (DNN) to replace different parts of traditional pipelines such as segmentation \cite{jalilian2017iris, fcn2016liu, irisdensenet} and representation \cite{zhao2017fcn, deepirisnet}. In particular, using ground-truth datasets such as IRISSEG-EP \cite{hofbauer2014ground}, recent works on fully convolutional network (FCN) based iris segmentation have shown promising improvements in robustness and accuracy.

Despite the improvements in segmentation accuracy with FCNs, existing studies focus solely on segmentation accuracy without evaluating the impacts of the models on end-to-end iris recognition systems. As evidenced in the work by Hofbauer \textit{et al.} \cite{hofbauer2016experimental}, segmentation accuracy alone may be insufficient when comparing multiple segmentation algorithms. In their study, they experiment with multiple iris recognition flows and demonstrate that segmentation algorithms with higher segmentation accuracy do not always lead to end-to-end flows with better recognition rate. Thus, when comparing multiple segmentation algorithms or models, it is helpful to evaluate each using the full iris recognition pipeline to select efficient models without sacrificing the overall system accuracy performance.

Existing works on FCN-based segmentation also lack evaluation of the model deployments on real HW/SW system such as embedded systems, which are popular targets for iris recognition applications. As such, the FCN architectures are designed without taking into account the computational overheads in deployment on resource-constraint systems. Instead, the narrow focus on segmentation accuracy also leads to FCN-based designs that are extremely computationally intensive. These models can consist of a large number of layers and parameters and require billions of floating-point operations for each input making them unsuitable for embedded systems. To address the current shortfalls, we propose in this work several contributions, which are summarized as follows:

\begin{itemize}
\item We propose an end-to-end iris recognition pipeline with FCN-based segmentation. To the best of our knowledge, we are the first to demonstrate a complete iris recognition flow using FCN-based segmentation. In order to construct this pipeline, we propose an accurate contour fitting algorithm which computes center points and radii of the pupil and limbic boundaries from the FCN segmented mask. The complete flow consists of an FCN-based segmentation, a contour fitting module, followed by Daugman normalization and encoding \cite{daugman2009iris}. Compared to previous works, our flow sets a new state-of-the-art recognition rate on the two datasets evaluated while incurring significantly less hardware resources.
\item The FCN-based segmentation portion is identified as the major bottleneck in our iris recognition pipeline. With this observation, we propose a three-step SW/HW co-design methodology to obtain a resource-efficient and accurate FCN model suitable for embedded platforms. Our technique consists of FCN architectural exploration, precision quantization using dynamic fixed-point format, and hardware acceleration.
\item First, in architectural exploration, we propose and evaluate a large number of FCN architectures and demonstrate that small decrease in segmentation accuracy can be traded off for an orders-of-magnitude reduction in overall computational complexities. Using the end-to-end flow, we highlight the importance of evaluating the impacts of various FCN architectures using overall recognition rates rather than just segmentation accuracy.
\item As a second step, we further reduce hardware complexities of the models by introducing quantization to 8-bit dynamic fixed-point for both weights and activations in the FCN models.
\item Next, we propose a novel custom, dynamic fixed-point based hardware accelerator design for the models. To compare with the floating-point format, we also synthesize a floating-point version of the accelerator.
\item Finally, we provide a hardware design space exploration and comparisons though implementation of the flow using various hardware configurations and precisions, namely CPU, CPU+Accelerator on FPGA, and CPU+GPU.
\end{itemize}

The rest of the paper is organized as follows. In Section~\ref{sec:background}, we provide a background of conventional and FCN-based iris segmentation and discussions of related work. In Section \ref{sec:methodology}, we describe in more details of our proposed iris recognition flow. Section~\ref{sec:swhw}, we discuss our resource-efficient SW/HW co-design methodology and hardware accelerator implementation. We discuss our experimental setup, as well as our experimental results in Section \ref{sec:experiment}. Finally, Section \ref{sec:conclusion} provides the final discussions and concludes the paper.

\begin{figure*}[tb!]
\centering
\includegraphics[scale=0.6, trim=2cm 8.1cm 2.75cm 8.1cm,clip]{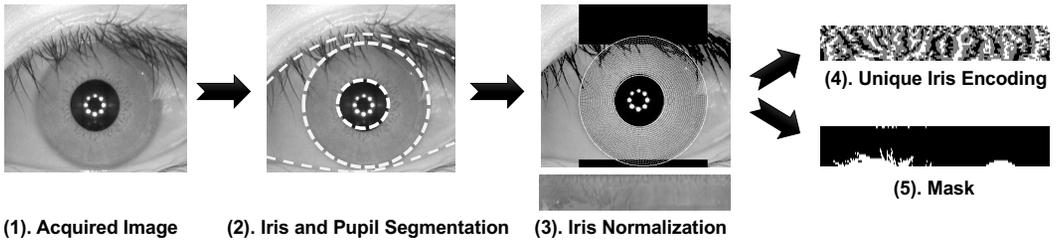}
\caption{Typical processing pipeline for iris recognition applications based on Daugman \cite{daugman2009iris}.}
\label{fig:pipeline}
\vspace{-0.15in}
\end{figure*}

\section{Background and Related Works} \label{sec:background}

In order to capture the unique features from each individual's irises and construct their corresponding signatures, the iris recognition pipeline typically consists of multiple stages as shown in Figure \ref{fig:pipeline}. First, the iris image is captured using a camera, often with near-infrared sensitivity. The input image is then preprocessed for specular reflections removal and contrast enhancement. Next, a segmentation step is applied to detect the pupil, iris and eyelids boundaries. The segmented iris region is then converted into its polar coordinate form in the normalization step. Finally, a wavelet transform is applied to encode the polar coordinate array into bitstream, which represents the unique signature of the iris \cite{daugman2009iris}. Each encoding is accompanied by a mask bit stream that gives encoding bits corresponding to none-iris areas such as those occluded by the eyelids or glare reflection. In this pipeline, the most computationally demanding portions are the preprocessing and iris segmentation \cite{lopez2011hardware, hashemi2018approximate}. In optimizing the pipeline, it is thus most beneficial to target these first few steps, which is the focus of this work.

\subsection{Traditional Iris Segmentation Methodologies}
Accurate iris segmentation has been among the most popular and challenging areas in iris recognition. One of the most widely adopted segmentation algorithms was proposed by Daugman \cite{daugman2009iris} using the integrodifferential operator. In this algorithm, the iris center point is located by searching through local-minimum intensity pixels throughout the image in a coarse-to-fine strategy. At each candidate pixel, a circular integrodifferential operator is applied while allowing the radius to change from a minimum to a maximum radius. This radius range is predetermined for the dataset to contain the limbic boundary. After all the candidate pixels are evaluated, the pixel location with the maximum in the blurred partial derivative with respect to the increasing radius is used in a fine-grain search. Here, integrodifferential operator is applied to all pixels in a small window surrounding the candidate pixels, which results in a single iris center point with radius, $r$. Once the iris radius and center points are determined, a similar step is used to search a small area around the iris center point for the pupil centers. Here, the radius range is allowed to vary from 0.1 to 0.8 of the computed iris radius. The integrodifferential operator is also used to determine the elliptical boundaries of the lower and upper eyelids.

Another popular technique used in many segmentation algorithms is circular Hough Transform \cite{wildes1994, kong2001accurate, ma2002iris, tisse2002person}. Typically, the Hough Transform operates on an edge map constructed from the input image. The main computation can be written as:
$$(x-x_i)^2 + (y-y_i)^2 = r^2$$
where $x_i$ and $y_i$ are the center coordinates, and $r$ is the circle radius. Similar to integrodifferential operator, the circle radius range for the iris and pupil boundaries are predetermined. A maximum in the Hough space corresponds to a most likely circle at radius $r$. The operator is used to compute two circles for the limbic and pupil boundaries. Since the iris region is often partially occluded by the top and bottom eyelids, two parabolic curves are used to approximate their boundaries.

The assumption of circular or elliptical limbic and pupil boundaries in the segmentation algorithms discussed can be challenging in some cases. For this reason, active contour-based segmentation algorithms were introduced to locate the true boundaries of the iris and pupil \cite{daugman2007new, shah2009iris,abdullah2017robust}. Since the segmentation output of active contour can assume any shapes, Abdullah \textit{et al.} \cite{abdullah2017robust} proposed a new noncircular iris normalization technique to unwrap the segmentation region. Gangwar \textit{et al.} \cite{irisseg} proposed a technique based on adaptive filtering and thresholding. Zhao and Kumar \cite{zhao2015tvm} proposed a total variation model to segment visible and infrared images under relaxed constraints.

\subsection{Fully Convolutional Networks for Iris Segmentation}

The challenges with traditional iris segmentation methods stem from the fact that the algorithms tend to be reliant on hand-crafted feature extractions and careful parameter tuning such as pre-computed radii ranges for the limbic and pupil boundaries. They can also be highly dependent on certain image intensity profiles and pre-processing steps to function correctly. In addition, separate models are typically deployed to detect the eyelids and iris regions.

\begin{figure}[t]
\centering
\includegraphics[scale=0.5, trim=6.2cm 5.5cm 6.2cm 4.5cm,clip]{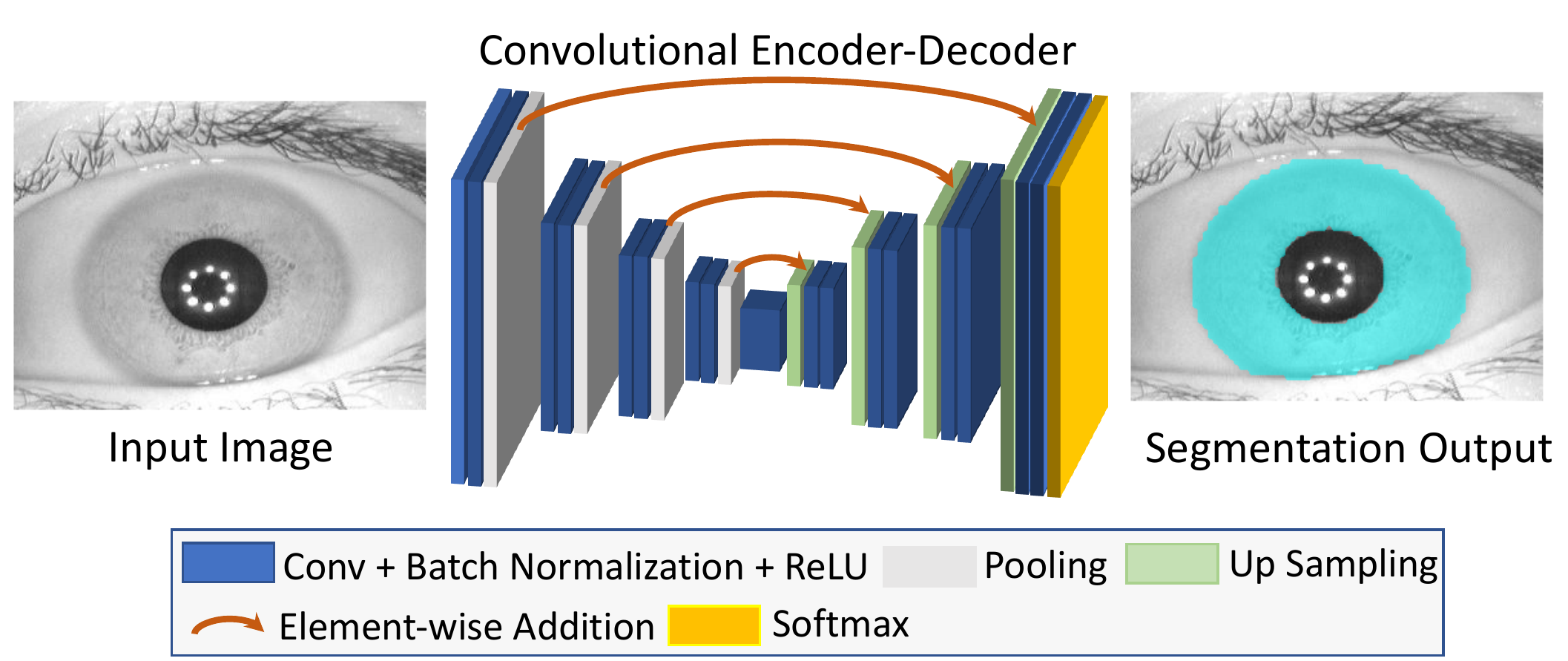}
\vspace{-0.1in}
\caption{Architecture for Encoder-Decoder Fully Convolution Networks with skip connections for semantic segmentation.}
\label{fig:fcn}
\vspace{-0.1in}
\end{figure}

With the recent advances in deep learning-based semantic segmentation, FCN-based iris segmentation methodologies have been proposed to solve the challenges facing conventional methods \cite{jalilian2017iris, fcn2016liu, irisdensenet, jalilian2017domain}. Similar to successful architectures used in other semantic segmentation problems such as SegNet \cite{segnet} and U-Net \cite{unet}, the state-of-the-art FCN models employed in iris segmentation typically has the form of encoder-decoder format as shown in Figure \ref{fig:fcn}. This architecture allows for pixel-wise labelling which conveniently produces output of the same dimensions as the input.

The success of the FCN models stem from their ability to learn and extract increasingly abstract features from the inputs. On the encoder side, it is observed that the hierarchical arrangement of convolutional layers allows earlier layers to learn lower-level features such as edges while latter layers learn more abstract, high-level concepts from the inputs. The underlying computation of each layer can summarized as convolution operations followed by a non-linear function such as Rectified Linear Unit (ReLU). The operation can be formalized as $$B_{i,j}=f(b+\sum_{m}\sum_{n}\sum_{k}(\textbf{A}_{i+m,j+n,k}\cdot\textbf{W}_{m,n,k}))$$
where $\textbf{A}$, $\textbf{W}$, and $b$ are the input tensor, kernel weight matrix, and a scalar bias respectively, and $f()$ is a non-linear function. A subset of the layers is also followed by a subsampling operation, which reduces the spatial dimension of the input allowing the model to be translation-invariant. On the decoder side, the low-resolution feature maps outputted by the encoder are upsampled using successions of transposed convolution layers to produce labeling prediction for each pixel in the original input image. 

\subsection{Metrics for Iris Segmentation Accuracy}
In order to evaluate segmentation algorithms, there exists multiple ways to compute the segmentation accuracy. A widely accepted metric in iris recognition is the $\mathcal{F}$-measure, which is aimed at optimizing the precision and recall performance of the segmentation output \cite{IR}. The resulting mask from a segmentation operation can be categorized into four different groups: true positive ($TP$), false positive ($FP$), true negative ($TN$) and false negative ($FN$). $TP$ and $TN$ represent fraction of pixels which were classified correctly as iris and none-iris respectively with respect to the ground truth segmentation. On the other hand, $FP$ and $FN$ correspond to those which are incorrectly classified as iris and none-iris. For a dataset with $N$ images, the precision is then defined as
$$\mathcal{P} \coloneqq \frac{1}{N}\sum_{i=1}^{N}\frac{TP_i}{TP_i+FP_i},$$
and recall is defined as
$$\mathcal{R} \coloneqq \frac{1}{N}\sum_{i=1}^{N}\frac{TP_i}{TP_i+FN_i}.$$
$\mathcal{P}$ measures the fraction of predicted iris pixels that is correct while $\mathcal{R}$ measures the fraction of iris pixels in the ground truth correctly identified or retrieved. $\mathcal{F}$ is then computed by taking the harmonic mean of $\mathcal{R}$ and $\mathcal{P}$:
$$\mathcal{F}\coloneqq \frac{1}{N}\sum_{i=1}^{N}\frac{2\mathcal{R}_i\mathcal{P}_i}{\mathcal{R}_i+\mathcal{P}_i}.$$

In iris recognition, other segmentation accuracy metrics also exist such as the Noisy Iris Challenge Evaluation - Part I \cite{nice}, where segmentation errors for a dataset of $N$ images, with $c\times r$ dimension, is defined as
$$E^1\coloneqq \frac{1}{N}\sum_{i=1}^{N}\left(\frac{1}{c\times r}\sum_{j=1}^{c\times r}O(j)\otimes C(j)\right).$$
Here, $O(j)$ and $C(j)$ are the pixels from the predicted outputs and ground truth masks respectively, and $\otimes$ is the XOR operator. A second error measure is also introduced which aims to compensate for the a priori probability disproportions between the iris and non-iris pixels in the input images:
$$E^2\coloneqq \frac{1}{2N}\sum_{i=1}^{N}(FP_i+FN_i).$$
In our work, we mainly utilize the $\mathcal{F}$-measure and also report the Precision and Recall performance. The $E^1$ and $E^2$ error rates can also be considered, but they would not affect our FCN optimization.

\section{Proposed FCN-based Iris Recognition Processing Pipeline} \label{sec:methodology}

\begin{figure}[tb]
\centering
\begin{tikzpicture}[
    node distance=5mm and 10mm,
box/.style = {draw, minimum height=12mm, align=center, rounded corners},
sy+/.style = {yshift= 0mm},
every edge quotes/.style = {align=center}
                        ]
\node (n1) [box]             {\textbf{Acquired}\\\textbf{Image}}; 
\node (n2) [box,right=of n1] {\textbf{FCN}\\\textbf{Inference}};
\node (n3) [box,right=of n2] {\textbf{Contour}\\\textbf{Fitting}};
\node (n4) [box,right=of n3] {\textbf{Normal-}\\\textbf{ization}};
\node (n5) [box,right=of n4] {\textbf{Encoding}};
\draw[thick,-Triangle]  
    ([sy+] n1.east) to ([sy+] n2.west);
\draw[thick,-Triangle]  
    ([sy+] n2.east) to ([sy+] n3.west);
\draw[thick,-Triangle]  
    ([sy+] n3.east) to ([sy+] n4.west);
\draw[thick,-Triangle]  
    ([sy+] n4.east) to ([sy+] n5.west);
\end{tikzpicture}
\caption{Proposed FCN-based iris recognition processing pipeline.}
\label{fig:fcnpipeline}
\vspace{-0.2in}
\end{figure}
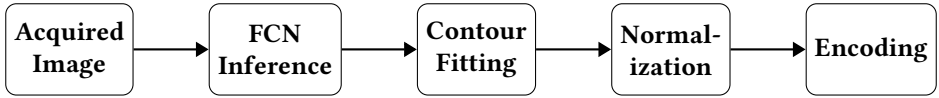

Traditional iris recognition pipelines consist of multiple computation stages for image pre-processing, segmentation, normalization, and encoding as depicted in Figure \ref{fig:pipeline}. In our flow, the segmentation is performed using an FCN model, which allows the pre-processing stage to be eliminated. For normalization and encoding, we employ the well-known rubber sheet model and 1-D log Gabor filter from Daugman \cite{daugman1993}. In order to connect the FCN segmentation output with the normalization stage, we propose a contour fitting routine, which will be described next. Figure~\ref{fig:fcnpipeline} illustrates our FCN-based processing pipeline which consists of FCN inference, contour fitting, normalization, and encoding.\\

\subsection{Proposed Contour Fitting Algorithm} \label{ssec:contour}
\begin{figure}[tb]
\centering
\includegraphics[scale=0.5, trim=5.3cm 4.1cm 6cm 3cm,clip]{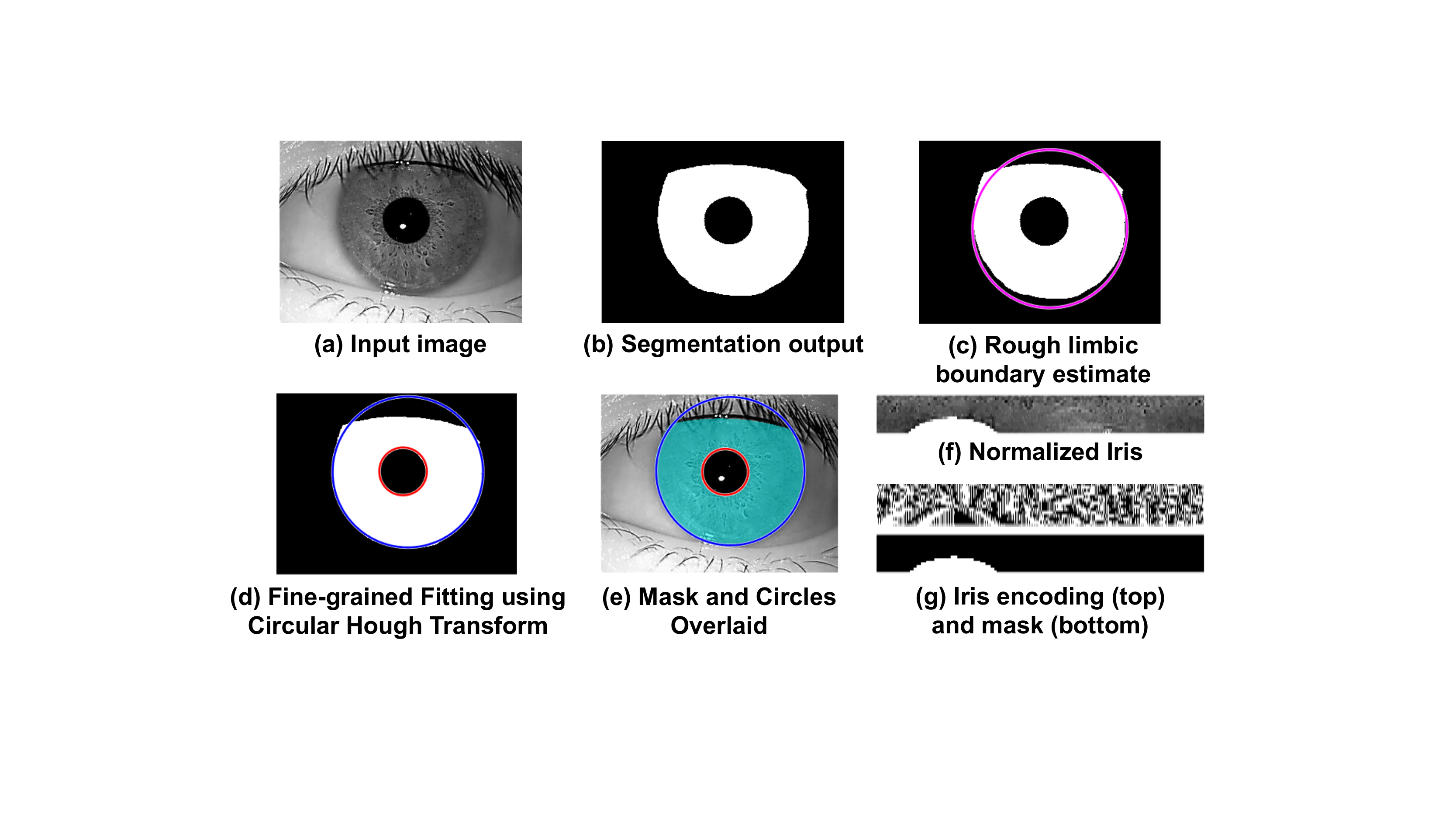}
\caption{Processing pipeline for contour fitting, normalization and encoding.}
\label{fig:contour}
\end{figure}

Daugman's rubber sheet model achieves iris 2D positional and size invariance due a new coordinate system created by the center points and radii of the iris and the pupil \cite{daugman2009iris}. With FCN-based segmentation, each output mask only identifies the pixels belonging to the iris and not the exact center coordinates or radii of the iris and the pupil. In order to extract this information, we develop a contour fitting routine as shown in Figure \ref{fig:contour}. Given a segmentation output mask, we first perform a rough estimate of iris center point and radius. This is done by analyzing the largest connected object in the image and computing its centroid, which is the rough iris center point. The iris radius is then approximated by taking the mean of the object's major and minor axis lengths.

Using the approximated center point and radius, we perform a more fine grained boundary fitting using the Circular Hough Transform (CHT) for circles with similar radii to the rough estimate. After obtaining the final iris radius ($r$) and center point ($x,y$), we search for the pupil using CHT for circles with radius range in the range [0.1$r$ 0.8$r$] and whose center points are within a region of interest (ROI) around $(x,y)$. We select this radius range because biologically, the pupil radius can be anywhere between 0.1 and 0.8 of the iris radius \cite{daugman1993}. The ROI allows for a less noisy and more computationally efficient localization of the pupil boundary.

\subsection{End-to-end Recognition Rates Evaluation} \label{ssec:end2end}
The contour fitting routine produces as output the information regarding the center coordinates and radii of the pupil and limbic boundaries. This result is passed on to the normalization step based on Daugman's rubber sheet model \cite{daugman2009iris}, which converts the iris region into a binary grid, 16$\times$256 in this work. A 1-D log Gabor filter is then used to extract features from the grid producing a 16$\times$256-bit encoding. A 16$\times$256-bit mask grid is also produced to identify useful and non-useful encoding bits. Note that, the Daugman normalization used in our current pipelines assumes circular limbic and pupilary boundaries. This assumption may not be suitable for some datasets such as those explored in \cite{daugman2007new} in which the recognition performance may be affected. However, it is a useful first order approximation, which can be built upon to better fit in those cases.

To determine whether there exists a match in the database, the hamming distance (HD) between the input encoding \{$encodingI$\} and every stored encoding \{$encodingS$\} is computed as follows:
\begin{equation}
    HD = \frac{||(encodingI \otimes encodingS) \cap maskI \cap maskS||}{||maskI\cap maskS ||},
    \label{eq:hamming}
\end{equation}
where \{$maskI$, $maskS$\} are the masks for input and stored encoding respectively. In our work, the HD is computed for different degrees of rotation in the range [-35$^{\circ}$, 35$^{\circ}$] between the two masks. From this, the smallest Hamming distance is recorded.

\section{Proposed SW/HW Co-design Strategy} \label{sec:swhw}

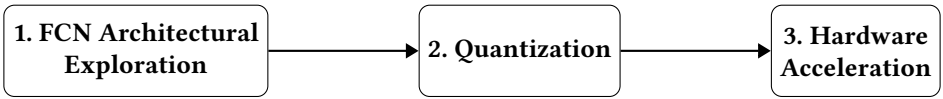
\begin{figure}[tb]
\centering
\begin{tikzpicture}[
    node distance=5mm and 20mm,
box/.style = {draw, minimum height=12mm, align=center, rounded corners},
sy+/.style = {yshift= 0mm},
every edge quotes/.style = {align=center}
                        ]
\node (n1) [box]             {\textbf{1. FCN Architectural}\\\textbf{Exploration}}; 
\node (n2) [box,right=of n1] {\textbf{2. Quantization}};
\node (n3) [box,right=of n2] {\textbf{3. Hardware}\\\textbf{Acceleration}};
\draw[thick,-Triangle]  
    ([sy+] n1.east) to ([sy+] n2.west);
\draw[thick,-Triangle]  
    ([sy+] n2.east) to ([sy+] n3.west);
\end{tikzpicture}
\caption{Proposed SW/HW co-design strategy to achieve efficient, accurate FCN model and fast inference.}
\label{fig:steps}
\vspace{-0.15in}
\end{figure}

We will show in Section \ref{ssec:implementation} that similar to most iris recognition pipelines, the segmentation step is the most compute intensive portion and takes up the majority of the overall processing time. In our flow, the segmentation runtime is mostly from FCN inference. Hence, we propose a three-step SW/HW co-design methodology, shown in Figure~\ref{fig:steps}, to reduce the hardware complexities for this processing stage while maintaining high accuracy. Next, we discuss each step in the methodology in more details.\\

\subsection{Fully Convolutional Network Architecture Exploration} \label{ssec:fcn_design}

In developing FCN models to perform iris segmentation, there are many choices for architectural parameters, each of which can lead to drastically different segmentation accuracy and computational complexities.
Generally, this design process uses empirical results from training and validating the models to refine the architectures. For this work, we explore network architectures similar to U-Net model proposed in \cite{unet}. We do not explore other model types such as DeepLab \cite{deeplab}, Segnet \cite{segnet}, and Long \textit{et al.} \cite{fcn} since they are targeted for more complex scenes with more training examples than our application.

In order to obtain the most efficient set of FCN architectures with good overall recognition performance, we first create a large pool of candidate FCN models with varying computational costs. Here, the \textit{computational cost} is defined as the number of arithmetic operations, which is the number of floating point operations (FLOPs) required per inference. We start by designing a baseline architecture as shown in Table \ref{tbl:fcnbase}. In this model, instead of using pooling layers to downsize the input, we employ strided convolution layers (convolutions with stride greater than 1). This has been shown to have no effect on the models' accuracy performance while offering reduced number of computations \cite{springenberg2014striving}. The larger models with more parameters, i.e. weights, tend to have the highest segmentation accuracy while requiring significant computational resources. However, the number of parameters must also be selected with care relative to the size of the available training data. Models with too many parameters on a small dataset can overfit and generalize poorly.

\begin{table*}[t]
\caption{Proposed baseline FCN architecture. Each convolution layer (CONV) is followed by Batch Normalization and ReLU activation layers. Transposed convolution layer (TCONV) is followed by ReLU activation layer. The arrows denote the shortcut connections, where the outputs of two layers are added together element-wise before passing to the next layer. The layers are grouped with a specific group number. The layers to below group number 4 are those belonging to the decoder side. While the encoder side are those to before and including group number 4. There always exists a shortcut connection between the last encoder-side CONV layer and decoder-side TCONV layer for groups with the same group number. For the baseline model, $N$ is set to be 16.}
\label{tbl:fcnbase}
\centering
\begin{tabular}{|c|ccc|}
\hline
Group \# & \textbf{Layer Type} & \textbf{Filter Size}/\textbf{Stride}/\textbf{Padding} & \textbf{Num. Outputs}\\
\hline
\hline
- & Image Input & -- & -- \\
\hline
0 & CONV\tikzmark{k0} & 3$\times$3/1/1 & N\\
\hline
\multirow{2}{*}{1} & CONV & 3$\times$3/2/1 & 2N\\
 & CONV\tikzmark{k1} & 3$\times$3/1/1 & 2N\\
\hline
\multirow{2}{*}{2} & CONV & 3$\times$3/2/1 & 2N\\
 & CONV\tikzmark{k2} & 3$\times$3/1/1 & 2N\\
\hline
\multirow{2}{*}{3} & CONV & 3$\times$3/2/1 & 2N\\
 & CONV\tikzmark{k3} & 3$\times$3/1/1 & 2N\\
\hline
\multirow{2}{*}{4} & CONV & 3$\times$3/2/1 & 2N\\
 & CONV & 3$\times$3/1/1 & 4N\\
\hline
\multirow{2}{*}{3} & T-CONV\tikzmark{kk3} & 4$\times$4/2/0 & 2N\\
 & CONV & 3$\times$3/1/1 & 2N\\
\hline
\multirow{2}{*}{2} & T-CONV\tikzmark{kk2} & 4$\times$4/2/0 & 2N\\
 & CONV & 3$\times$3/1/1 & 2N\\
\hline
\multirow{2}{*}{1} & T-CONV\tikzmark{kk1} & 4$\times$4/2/0 & 2N\\
 & CONV & 3$\times$3/1/1 & 2N\\
\hline
\multirow{2}{*}{0} & T-CONV\tikzmark{kk0} & 4$\times$4/2/0 & N\\
 & CONV & 1$\times$1/1/1 & 2\\
\hline
- & SOFTMAX & -- & 2\\
\hline
\end{tabular}
\begin{tikzpicture}[overlay, remember picture, yshift=.25\baselineskip, shorten >=.5pt, shorten <=.5pt]
\draw [->] ({pic cs:k0}) [bend left] to ({pic cs:kk0});
\draw [->] ({pic cs:k1}) [bend left] to ({pic cs:kk1});
\draw [->] ({pic cs:k2}) [bend left] to ({pic cs:kk2});
\draw [->] ({pic cs:k3}) [bend left] to ({pic cs:kk3});
\end{tikzpicture}
\end{table*}

With a baseline architecture designed, we iteratively construct different FCN variants by performing a grid search on a few architectural parameters. The choices of parameters are chosen such that they have significant impact on the computational complexities of the models. The three parameters are as follows:
\begin{itemize}
\item \textit{Input image scaling}: The spatial dimensions of the input iris image directly affect the number of computation required at each layer. While the original image resolution offers more detailed and fine features, segmentation using scaled-down version of the input could offer significant reduction in number of computation with limited effect on the segmentation accuracy. We explore three different scaling factors in this work, namely, 1 (original resolution), 0.5, and 0.25. For instance, a scaling factor of 0.5 means that the each spatial dimension of the input image is reduced by half.
\item \textit{Number of layers}: We explore FCN models with wide ranging number of layers for each dataset. The maximum number of layers explored is 18 as shown in Table \ref{tbl:fcnbase}. We obtain models with smaller number of layers by removing layers in groups. For instance, removing layers with group number 3 would results in model a 14-layer network. However, we set a strict constraint that the spatial dimensions of the smallest feature maps in the models, namely the outputs of group 4, are kept fixed at $\frac{1}{16}$ the original dataset resolution.
\item \textit{Number of feature maps/channels per layer}: This parameter is denoted by variable $N$ in Table~\ref{tbl:fcnbase} and quadratically impacts the computational complexity of each FCN layer. For efficiency, we limit the maximum number of output feature maps to be 64 in any layer. Starting from the baseline architecture, we experiment with four different values for $N$, which are \{4, 8, 12, 16\}.
\end{itemize}
However, several architectural choices are kept constant across all the models. For instance, the filter size of all convolution layers are also kept fixed at 3$\times$3 except for the last convolution layer, which is 1$\times$1. The size is 4$\times$4 for all transposed convolution layers. None-strided convolution layers are padded to keep the spatial dimensions of the output feature maps the same as their inputs.

Each candidate model is trained using the backpropagation algorithm with stochastic gradient descent (SGD) and momentum weight updates:
$$\Delta\textbf{W}_{t+1} = \beta\Delta\textbf{W}_{t}- \eta \nabla \mathcal{L}(\textbf{W})$$
$$\textbf{W}_{t+1} = \textbf{W}_{t}+\Delta\textbf{W}_{t+1} $$
where $\beta$ and $\eta$ are the momentum and learning rate respectively. For loss function $\mathcal{L}(\textbf{W})$, we use cross entropy loss where there are two output classes, iris and non-iris for each pixel. This loss can be written as:
$$\mathcal{L}(\textbf{W}) = -\frac{1}{c\times r}\sum_{i=1}^{c\times r} (y_i\log{p_i} + (1-y_i)\log(1-p_i)),$$
where $y_i\in\{0,1\}$ and $p_i\in[0,1]$ are the ground truth and predicted label for each pixel respectively. This loss function works well in case where the number of pixels in each class is roughly equal. In reality, most images captured for iris recognition contain much smaller iris area compared to non-iris. Thus, we introduce additional parameter to compensate for the disproportionality of the two classes a priori probabilities as:
$$\mathcal{L}(\textbf{W}) = -\frac{1}{c\times r}\sum_{i=1}^{c\times r} ((1-\alpha)(y_i\log{p_i}) + \alpha(1-y_i)\log(1-p_i)),$$
where $\alpha \in [0,1]$ is ratio of iris to non-iris area and precomputed from the training set.\\

\begin{figure}[tb!]
\centering
\includegraphics[scale=0.425, trim=0.05cm 0.35cm 0.2cm 0.42cm,clip]{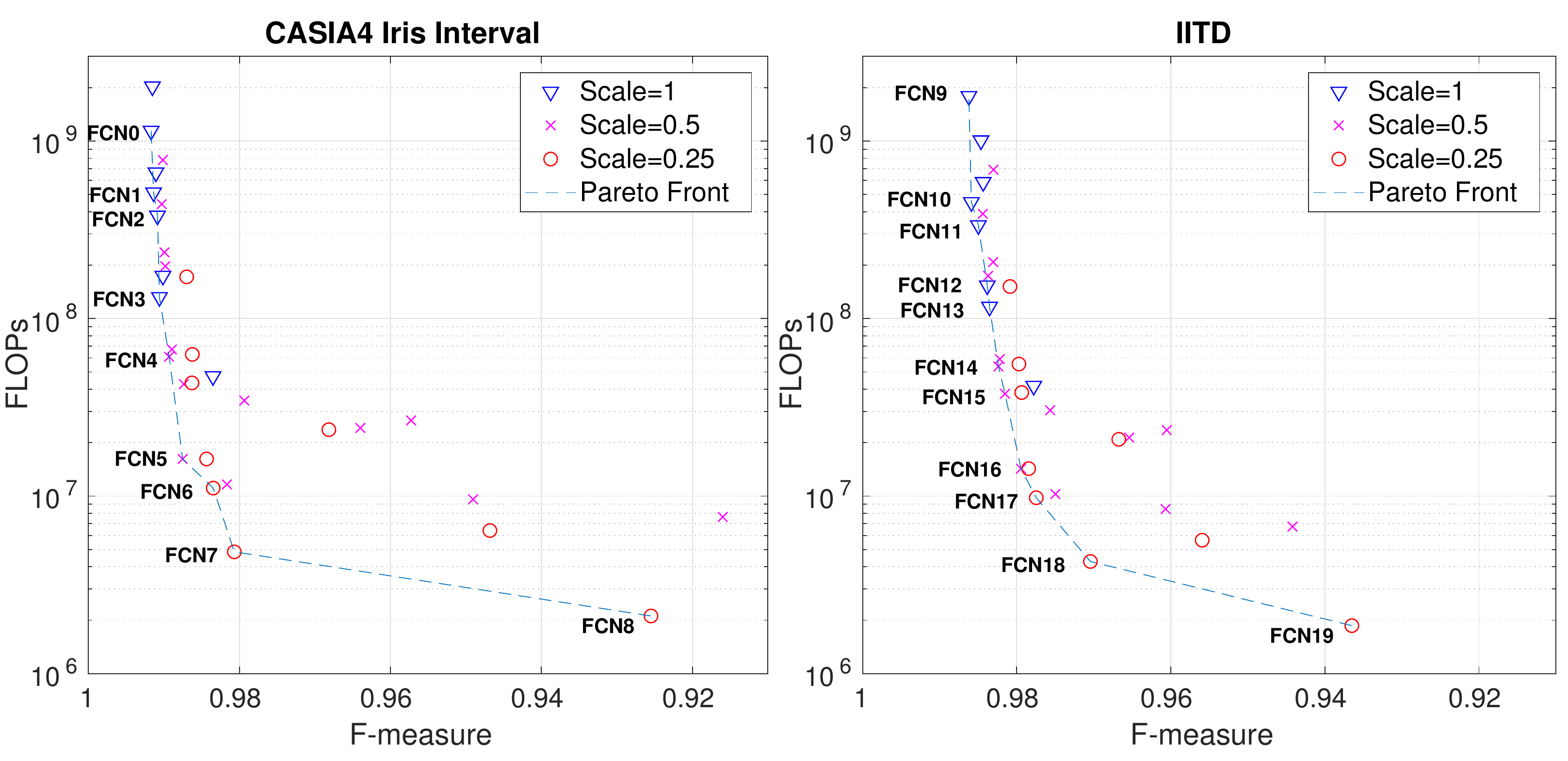}
\caption{$\mathcal{F}$-measure segmentation accuracy and computational complexity of candidate FCN models on CASIA Iris Interval V4 and IITD datasets. The scales refer to the ratio of the model input dimensions to the original image resolutions from the datasets. Smaller resolution inputs can significantly reduce the computational complexity of the models. We label models which make up the Pareto fronts as FCN0-FCN8 for CASIA4 and FCN9-FCN19 for IITD.}
\label{fig:candidates}
\vspace{-0.15in}
\end{figure}

\noindent\textbf{Segmentation accuracy evaluations:} We evaluate two well-known datasets in this work, namely CASIA Interval V4 \cite{casiav4} and IITD \cite{iitd}. Figure \ref{fig:candidates} shows the $\mathcal{F}$-measure performance and computational complexity, defined as the number of FLOPs required per inference, of candidate FCN models evaluated. For each dataset, the models were trained on a training set, and the reported $\mathcal{F}$-measures in Figure \ref{fig:candidates} are obtained using a disjoint test set. The training and validation sets are 80\% of the original dataset with the remaining 20\% for the test set. For models using scaled-down images, each input is first downsized according to the scale factor. The output segmentation mask is then resized back to the original resolution before the $\mathcal{F}$-measure is computed. We use the nearest-neighbor approach for the both resizing operations. Note that in our architectural explorations, we train separate networks for the two datasets for fair comparisons with previous works. This does not limit the applicability of our models as techniques such as domain adaptation \cite{jalilian2017domain} can be applied for new unseen datasets.

\begin{table*}[!tb]
\caption{Architectural descriptions for FCN models which form the Pareto fronts in Figure~\ref{fig:candidates} for CASIA Interval V4 and IITD datasets. Scaling denotes the input image scaling, and $N$ determines the number of channels as in Table~\ref{tbl:fcnbase}. The architectures are shown as a list of group number, each of which corresponds the various layers shown in Table~\ref{tbl:fcnbase}. The layers to the right of group number 4 are those belonging to the decoder side. While the encoder side are those to the left of and including group number 4. Similar to Table \ref{tbl:fcnbase}, there exists shortcut connections between the last encoder-side CONV layers and decoder-side TCONV layers for groups with the same group number, which are not shown here.}
\label{tbl:fcn}
\centering
\begin{tabular}{|c|cc|c|cc|}
\hline
\multirow{2}{*}{\textbf{Model}} & \multicolumn{2}{|c|}{\textbf{CASIA Interval V4}} & \multirow{2}{*}{\textbf{Model}} & \multicolumn{2}{c|}{\textbf{IITD}}\\
\cline{2-3}
\cline{5-6}
& \textbf{Scaling/$N$} & \textbf{Architecture} & & \textbf{Scaling/$N$} & \textbf{Architecture}\\
\hline
\hline
FCN0 & 1/12 & 0-1-2-3-4-3-2-1-0 &  FCN9 & 1/16 & 0-1-2-3-4-3-2-1-0 \\
FCN1 & 1/8 & 0-1-2-3-4-3-2-1-0 &   FCN10 & 1/8 & 0-1-2-3-4-3-2-1-0 \\
FCN2 & 1/12 & 0-1-2-4-2-1-0 &      FCN11 & 1/6 & 0-1-2-3-4-3-2-1-0 \\
FCN3 & 1/4 & 0-1-2-3-4-3-2-1-0 &   FCN12 & 1/4 & 0-1-2-3-4-3-2-1-0 \\
FCN4 & 0.5/8 & 0-1-2-4-2-1-0 &     FCN13 & 1/4 & 0-1-2-4-2-1-0 \\
FCN5 & 0.5/4 & 0-1-2-4-2-1-0 &     FCN14 & 0.5/8 & 0-1-2-4-2-1-0 \\
FCN6 & 1/4 & 0-1-2-4-2-1-0 &       FCN15 & 0.5/4 & 0-1-2-4-2-1-0 \\
FCN7 & 0.25/4 & 0-1-4-1-0 &        FCN16 & 0.25/8 & 0-1-4-1-0 \\
FCN8 & 0.25/4 & 0-4-0 &            FCN17 & 0.25/4 & 0-1-4-1-0 \\
 &  &  &                           FCN18 & 0.25/8 & 0-4-0 \\
 &  &  &                           FCN19 & 0.25/4 & 0-4-0 \\
\hline
\end{tabular}
\vspace{-0.1in}
\end{table*}

As illustrated in Figure \ref{fig:candidates}, different $\mathcal{F}$-measures can result in drastic difference in FCN computational complexities. For the two datasets, our architectural explorations result in models with three orders of magnitude range in complexity, between 0.002 and 2 GFLOPs. The results also show that models using input size closer to the original resolution tend to perform slightly better, however, they are significantly more complex computationally than the lower resolution counterpart. In addition, for each input size, the different architectural choices can lead of orders of magnitude difference in number of computations and segmentation accuracy. For both datasets, the accuracy performance for models using different input scaling saturates at different point beyond which small additional accuracy improvement require orders of magnitude increase in complexity. This saturation behavior is also observed when all scaling factors are combined. We provide architectural descriptions of each model from the Pareto fronts (FCN0-FCN8 and FCN9-FCN19) for the two datasets in Table~\ref{tbl:fcn}.

To compare the efficiency and segmentation performance of our models to previous works, we also evaluate each model using the full dataset. Table \ref{tbl:comparison} shows the results from our best-performing model and those from previous works. The segmentation accuracy of other works reported in the table are obtained from IrisSeg \cite{irisseg} and IrisDenseNet (IDN) \cite{irisdensenet}. Previously, IrisSeg achieved better segmentation accuracy performance in comparison to other none-FCN segmentation methods such as GST \cite{gst}, Osiris \cite{osiris}, Masek \cite{masek}, WAHET \cite{wahet}, and CAHT \cite{caht}. This result was outperformed by FCN-based segmentation method proposed by IDN from Arsalan \textit{et al.} \cite{irisdensenet}. In comparison to IDN model, which requires more than 100 GFLOPs per inference, both of our FCN architectures need less than 2 GFLOPs as shown in Table \ref{tbl:fcn}, which is 50$\times$ more efficient. This large difference in computational overhead can be attributed to the fact that our network architectures are significantly shallower with far fewer number of feature maps per layer. In addition, our models utilize few shortcut connections instead of the costly dense connectivity.

\begin{table}[!tb]
\caption{Segmentation Accuracy Comparison to Previous Works}
\label{tbl:comparison}
\centering
\begin{tabular}{|c|ccccccc|}
\hline
\multirow{2}{*}{\textbf{DB}} & \multirow{2}{*}{\textbf{Method}} & \multicolumn{2}{c}{\textbf{R}} & \multicolumn{2}{c}{\textbf{P}} & \multicolumn{2}{c|}{\textbf{F}}\\
\cline{3-8}
& & $\mu$ & $\sigma$ & $\mu$ & $\sigma$ & $\mu$ & $\sigma$\\
\hline
\multirow{8}{*}{\rotatebox[origin=c]{90}{\textbf{CASIA4 Interval}}} & GST \cite{gst} & 85.19 & 18 & 89.91 & 7.37 & 86.16 & 11.53\\
& Osiris \cite{osiris} & 97.32 & 7.93 & 93.03 & 4.95 & 89.85 & 5.47\\
& WAHET \cite{wahet} & 94.72 & 9.01 & 85.44 & 9.67 & 89.13 & 8.39\\
& CAHT \cite{caht} & 97.68 & 4.56 & 82.89 & 9.95 & 89.27 & 6.67\\
& Masek \cite{masek} & 88.46 & 11.52 & 89.00 & 6.31 & 88.30 & 7.99\\
& IrisSeg \cite{irisseg} & 94.26 & 4.18 & 92.15 & 3.34 & 93.10 & 2.65\\
& IDN \cite{irisdensenet} & 97.10 & 2.12 & 98.10 & 1.07 & 97.58 & 0.99\\
& Our FCN0 & \textbf{99.41} & \textbf{0.40} & \textbf{98.93} & \textbf{0.75} & \textbf{99.17} & \textbf{0.40}\\
\hline
\multirow{8}{*}{\rotatebox[origin=c]{90}{\textbf{IITD}}} & GST \cite{gst} & 90.06 & 16.65 & 85.86 & 10.46 & 86.60 & 11.87\\
& Osiris \cite{osiris} & 94.06 & 6.43 & 91.01 & 7.61 & 92.23 & 5.80\\
& WAHET \cite{wahet} & 97.43 & 8.12 & 79.42 & 12.41 & 87.02 & 9.72\\
& CAHT \cite{caht} & 96.80 & 11.20 & 78.87 & 13.25 & 86.28 & 11.39\\
& Masek \cite{masek} & 82.23 & 18.74 & 90.45 & 11.85 & 85.30 & 15.39\\
& IrisSeg \cite{irisseg}  & 95.33 & 4.58 & 93.70 & 5.33 & 94.37 & 3.88\\
& IDN \cite{irisdensenet} & 98.00 & 1.56 & 97.16 & 1.40 & 97.56 & 1.04\\
& Our FCN9 & \textbf{98.92} & \textbf{0.87} & \textbf{98.33} & \textbf{1.13} & \textbf{98.62} & \textbf{0.65}\\
\hline
\end{tabular}
\vspace{-0.1in}
\end{table}

\subsection{Quantization to Dynamic Fixed-Point} \label{ssec:iris_quantization}
As demonstrated by Hashemi {\it et al.} \cite{hashemi2018approximate} and Tann {\it et al.} \cite{tann2017hardware}, reducing the data precision in DNNs can significantly lower the computational overheads of the models. With the Pareto front models identified in Figure~\ref{fig:candidates}, we co-design their data precision such that they can be run using lower-cost computational units on the targeted hardware platform. Since quantization is a time-consuming process, we do not target other models which are not on the Pareto fronts.

The numerical ranges of the weights and activations in DNN models can vary drastically between different layers. Previous works ~\cite{hashemi2017understanding} have shown that even quantizing the weights and activations to a 16-bit uniform fixed-point format significantly degrades the accuracy of models in comparison to the original floating-point representation. In order to represent these different ranges using a small number of bits, we propose to quantize the FCN models to dynamic fixed-point (DFP) \cite{courbariaux2014training} for both the weights and activations. Within a layer, DFP behaves exactly like a normal fixed-point format. However, the radix location is allowed to vary between different layers for DFP. In this format, each layer in the FCN models is represented by five hyperparameters, namely ($w_{bw}$, $a_{bw}$, $w_{fl}$, $a_{in}$, $a_{out}$), for bitwidths of the weights and activations/feature maps, and fractional lengths of the weights, input feature maps, and output feature maps respectively. We fix the bitwidths of both weights and activations of all the layers to be 8 bits.

In order to determine the proper fractional lengths for the weights and feature maps of each layer, we first perform profiling for the weights and activations of the trained floating-point models. For the weights, we select layer-wise fractional lengths such that no overflow exists during the quantization. For the activations or feature, the profiling is done by using a randomly selected subset of training data to perform forward passes with the models. During this inference process, we record the largest activation for each layer. Similar to the weights, we then select layer-wise fractional lengths such that there is no overflow. With these hyperparameters in place, we then quantize the floating models to DFP by employing similar procedure to Hashemi {\it et al.} \cite{hashemi2017understanding} using the straight-through estimator.

\subsection{Hardware Acceleration Iris Recognition Pipeline on Embedded SoC} \label{ssec:implementation}
So far, the majority of work on iris recognition focuses mostly on algorithmic designs such as segmentation and feature extraction. There exists only few studies on the system design and implementation aspect. Hashemi \textit{et al.} \cite{hashemi2018approximate} and L{\'o}pez \textit{et al.} \cite{lopez2011hardware} implemented full recognition pipelines on an embedded FPGA platform and showed that careful parameters optimization and software-hardware partitioning are required to achieve acceptable runtime. For iris recognition with FCN-based segmentation, existing studies so far are only concerned with achieving state-of-the-art segmentation accuracy without considerations for computational costs of the proposed designs. As such, full system analysis and implementation of these processing pipelines have not been demonstrated. In this section, we describe our implementation of the FCN-based iris recognition pipeline targeting an embedded FPGA SoC. We provide analysis of the system runtimes and bottlenecks. We also propose a hardware accelerator, which is able to achieve significant speedup computations relative to the onboard CPU core.

\subsubsection{Runtime Profiles for Iris Recognition Pipeline}
As an initial step, we implement the iris recognition pipeline in software running on the physical CPU core on the FPGA SoC. Our pipeline consists of four main modules, namely segmentation, contour fitting, normalization, and encoding. The segmentation step can be performed using different FCN models, which can lead to vastly different runtimes. On the other hand, the runtimes for the remaining three components stay approximately constant across different input images and FCN models. This is because the dimensions of the input and output images for these three modules are constant.

\begin{figure}[tb!]
\includegraphics[scale=0.6, trim=0.74cm 1cm 1.2cm 0.5cm,clip]{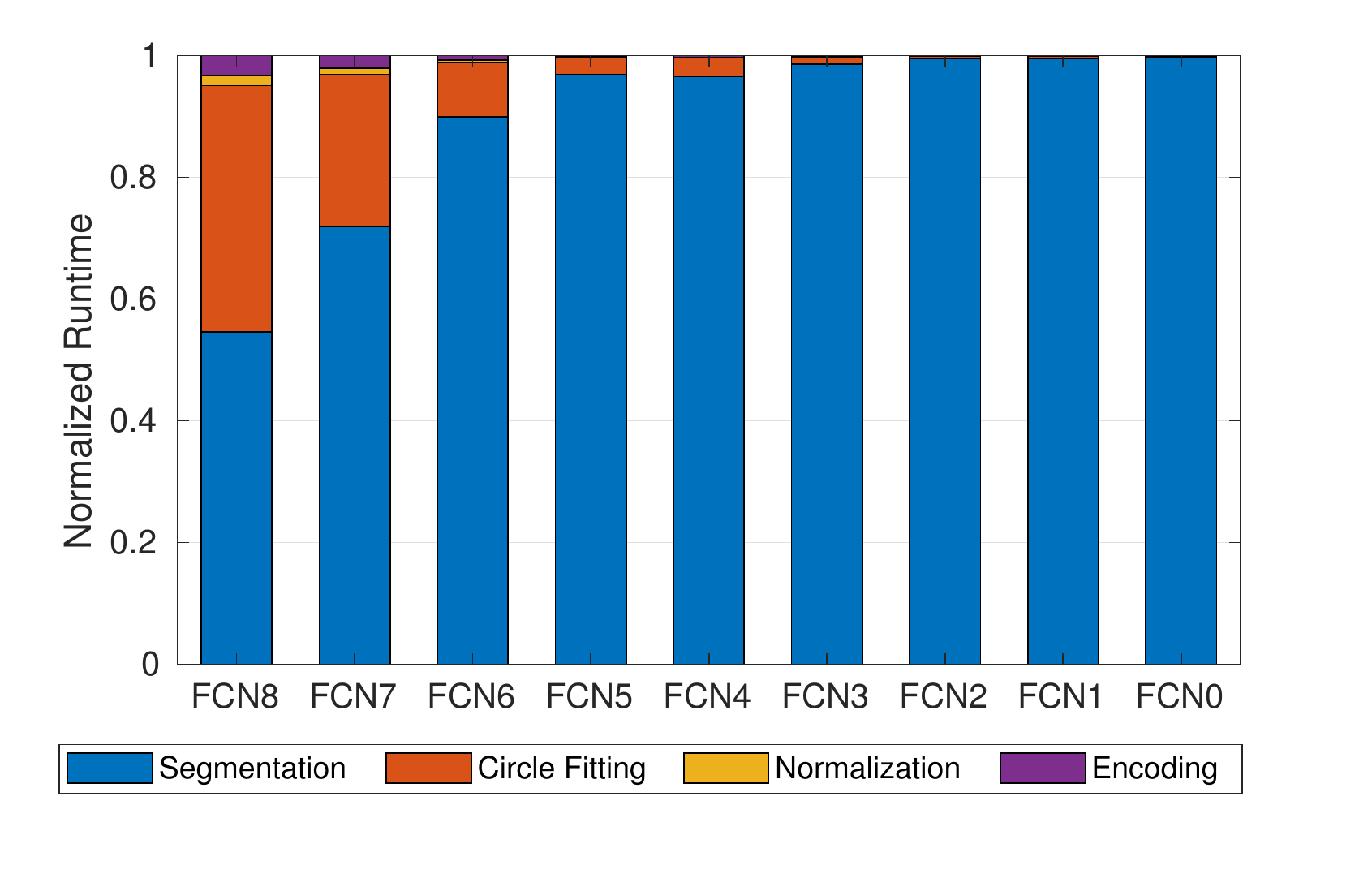}
\caption{FCN-based iris recognition pipeline runtime breakdown for FCN0--FCN8 models from CASIA Interval V4 Pareto front in Figure \ref{fig:candidates}. From left to right, the FCN models are arranged in increasing computational complexity.}
\label{fig:runtimebar}
\vspace{-0.15in}
\end{figure}

With this setup, we profile the runtime of the different components in the pipeline, which is shown in Figure \ref{fig:runtimebar}. Here, we observe that the majority of the runtime is spent in the segmentation stage. This is especially true for larger FCN models where segmentation takes up more than 95\% of the total runtime. Therefore, it is reasonable to focus our efforts on accelerating the segmentation component of the pipeline, which is essentially the inference process of the FCN model. To effectively speed up this operation, we explore next the runtime profiles for FCN model components.

\begin{table}[tb]
\renewcommand{\arraystretch}{1.3}
\caption{Runtime profile for FCN inference using the onboard CPU.}
\label{table:fcnprofile}
\centering
\begin{tabular}{|c|c|c|c|c|}
\hline
Function & Init & Im2Col & GEMM & Activation (ReLU)\\
\hline
Percentage & 1.31 & 10.58 & 80.77 & 7.34\\
\hline
\end{tabular}
\vspace{-0.1in}
\end{table}

\subsubsection{FCN Processing Components}

In this work, our FCN models are implemented and trained using the Darknet framework \cite{darknet}. Each model consists of multiple layers with different computational requirements, and each layer consists of multiple components as listed in Table \ref{table:fcnprofile}.
Here, the Init functions is responsible for ensuring that the output matrices are properly initialized and zeroed out. Note that Batch Normalization (BN) layers are used in training, but they are not shown here since the trained normalization parameters ($\mu, \sigma^2, \gamma, \beta$) can be folded into the network parameters in inference as such: 
$$\hat{\textbf{w}} = \gamma\cdot\textbf{w}/\sigma^2$$
$$\hat{\textbf{b}} = \gamma\cdot(\textbf{b}-\mu)/\sigma^2 + \beta$$
where \textbf{w} and \textbf{b} are the trained weights and biases of the preceding convolution layer. With this, the forward computation can be carried out using $\hat{\textbf{w}}$ and $\hat{\textbf{b}}$ without the BN layers. The Im2Col function 
is a standard operation which converts the input images/feature maps into column format. With this, the convolution operations can be carried out using a general matrix to matrix multiplication (GEMM) routine. For transposed convolution layer, a similar operation is used to convert column data to image instead. The GEMM unit is essentially responsible for the multiplication of two matrices, the weights and input feature maps. The results in Table \ref{table:fcnprofile} show that the GEMM unit is the most time consuming portion taking up more than 80\% of the module runtime. The remaining 20\% is spent mostly on Im2Col and activation function, which is the rectify linear unit in this case.


The resources on-board the SoC allow for multiple choices for accelerating the pipeline including parallelization and vectorization using embedded CPU cores and custom hardware accelerator on the programmable logic (PL) fabric. In comparison to the PL, parallelization and vectorization on the CPU offer limited number of arithmetic processing units; however, accelerators on the PL side can face challenges in the limited on-chip buffer and memory bandwidths. Thus, in order to efficiently utilize the available hardware resources, we leave the control logic and memory-access intensive component, Im2Col, in software and move computational intensive module, GEMM, to PL by synthesize a custom accelerator. For the activation function, we process it using the CPU core in parallel to the accelerator unit. Next, we describe in details our accelerator architecture.

\subsubsection{Hardware Accelerator Architecture}

\begin{figure*}[tb!]
\centering
\includegraphics[scale=0.57, trim=1.75cm 5.5cm 2cm 4.1cm,clip]{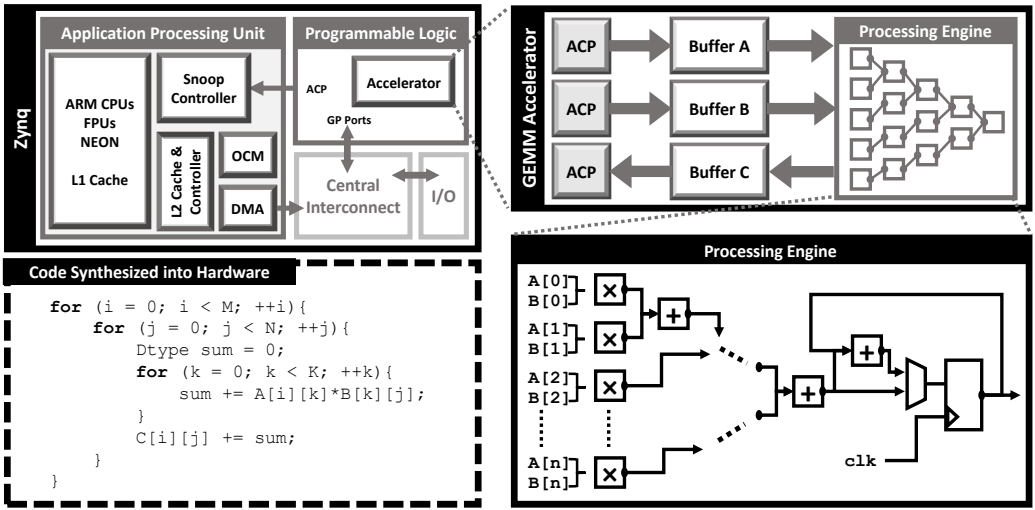}
\caption{Overall system integration and the hardware accelerator module for the GEMM unit. The code representing the operations of the hardware module is shown in the bottom left, where A and B are the multiplicant and multiplier matrices, and C is the resulting output matrix. For DFP version of the accelerator, A and B are 8-bit, and C is 16-bit. A, B and C are all 32-bit floats for the floating-point version. The accelerator module is connected to the Zynq Processor Unit via the Accelerator Coherency Port (ACP).}
\label{fig:hardware}
\vspace{-0.1in}
\end{figure*}

For FCN models, the GEMM operation is carried out in every layer between the weight and input feature matrices. The dimensions of the two matrices can be represented by a 3-tuple, (\textit{M, K, N}), where the weight matrix is $M\times K$, and the input features matrix is $K\times N$. The output feature matrix is then $M\times N$. Between different layers of an FCN model, (\textit{M, K, N}) vary significantly depending the on sizes and number of the input and output feature maps. An evidence of this can be observed in the our network architecture shown in Table \ref{tbl:fcn} for CASIA Interval V4. In this architecture, after Im2Col operation, the (\textit{M, K, N}) dimensions would be (16, 9, 76800) for Layer 1, where as for Layer 2, these dimensions become (32, 144, 19200). Among FCN models which use different input image scaling factors, these dimensional differences are even more drastic. As such, the accelerator unit must be able to accommodate these dimensional variations and maximize utilization across all the models explored.

Figure \ref{fig:hardware} shows the overall system integration and the architecture of the accelerator core. We implement tiling buffers for the weights (Buffer A), input features (Buffer B), and output features (Buffer C). The sizes of these buffers are selected based on the greatest common divisor among the models. For the candidate models in Figure \ref{fig:candidates}, these turn out to be 8$\times$9 for matrix A, 9$\times$224 for B, and finally 8$\times$224 for matrix C. Note that, since we do not target a specific model, the sizes for A, B, and C may not be optimal for any specific architecture. In final system deployment, such dimensions can be further optimized according to the chosen FCN model. We used Vivado High Level synthesis (HLS) to develop the GEMM accelerator, which is connected to external memory via an AXI4-Full interface to Accelerator Coherency Port (ACP). A DMA is used to communicate with ACP and fill the accelerator buffer. Here, we use the ARM CPUs as the control unit through a separate AXI-Lite connection. The CPU is responsible for preparing and feeding correct addresses in of the input and output matrices as well as sending the start signal. Once this start signal is received, the accelerator unit accesses the input matrices, performs computations and writes the output matrix to the designated address in the DDR RAM.

\begin{figure*}[tb]
\centering
\includegraphics[scale=0.5, trim=3.5cm 7cm 3.5cm 6cm,clip]{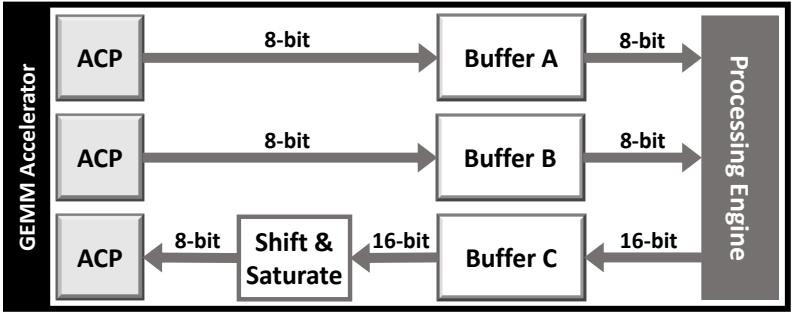}
\caption{A closer look at the data paths of the buffers in the DFP accelerator unit.}
\label{fig:datapath}
\vspace{-0.1in}
\end{figure*}

The accelerator in Figure \ref{fig:hardware} utilizes nine parallel floating-point multipliers each of which is connected to different banks of block RAM contain portions of input from matrices A and B. This matrix partitioning helps improve the throughput of the design. The output of the multipliers are then summed together using an adder tree consisting of 9 adders. If the output is a partial sum, it is written to buffer C for accumulation until completion before being written back to the DRAM. For the floating-point version, all the datapaths and buffers are 32-bit wide. For the DFP version, Figure~\ref{fig:datapath} provides a closer look at the datapaths. Since DFP representation may result in different radix-point location for the feature maps between different FCN layers, we need to shift the output results accordingly. Afterward, the output feature maps are converted to 8-bit and saturated if necessary.

\section{Experimental Results} \label{sec:experiment}
In this section, we discuss the segmentation and recognition performance of our proposed processing pipeline. We also report the runtime performance for the embedded FPGA implementation, speedup achieved using our hardware accelerator, and comparison to an embedded GPU.

\subsection{Experimental Setup}
All of our experiments are performed using two well-known and publicly available iris datasets, the CASIA Interval V4 \cite{casiav4} and IITD \cite{iitd}. Both datasets are captured using near-infrared range sensors and reflect real-world deployment conditions. The ground truth segmentation masks used in all of our experiments are obtained from IRISSEG-EP \cite{hofbauer2014ground}. We use segmentation from Operator A for CASIA Interval V4 dataset. For FCN training and deployment, we use the Darknet framework \cite{darknet}. We fully implement our processing pipeline on the ZedBoard with Xilinx Zynq 7020 FPGA SoC with 28nm process node and 512MB DDR3 memory. The chip contains two ARM Cortex A9 cores and programmable logic fabric with 85K logic cells and 4.9Mb block RAM. We also implement the flow on an embedded GPU, namely the NVIDIA Jetson TX1, for comparison. The TX1 SoC has 4 ARM Cortex A57 cores, 256 Maxwell CUDA Cores all with 20nm process node, and 4GB of LPDDR4 memory. For both systems, our iris recognition flow is run inside an embedded linux operating system.

\subsection{Recognition Performance Comparisons to Previous Works}

While isolated evaluation of FCN models using the segmentation accuracy can be helpful in narrowing down to the most efficient set of models, they are not a sufficient indicator of the true overall recognition performance. The true trade-off between FCN model computational complexity and recognition performance can only be analyzed using an end-to-end flow. That is each model must be evaluated based on performance metrics such as equal error rate (EER) and its receiver operating characteristics (ROC). Since end-to-end evaluation on all models explored is extremely time consuming, we select only the models from the Pareto fronts from Figure \ref{fig:candidates}, which represent the most efficient models across the segmentation accuracy levels. The models on the Pareto fronts are labeled FCN0--FCN8 and FCN9--FCN19 for CASIA Interval V4 and IITD datasets respectively. For each dataset, the labels are in decreasing order of computational complexity as well as segmentation accuracy.

To evaluate the recognition performance of each FCN model, we perform all possible combinations of intra-class, which are different instances of the same iris, and inter-class matchings. For CASIA Interval V4, this results in approximately 9K intra-class and 6.9M inter-class comparisons. For IITD, approximately 4.8K intra-class and 5M inter-class comparisons are performed. In each matching, the hamming distance (HD) is computed as described in Section~\ref{ssec:end2end}.

\begin{figure*}[tb!]
\begin{center}
\includegraphics[scale=0.45, trim=0.05cm 0.1cm 0.5cm 0.2cm,clip]{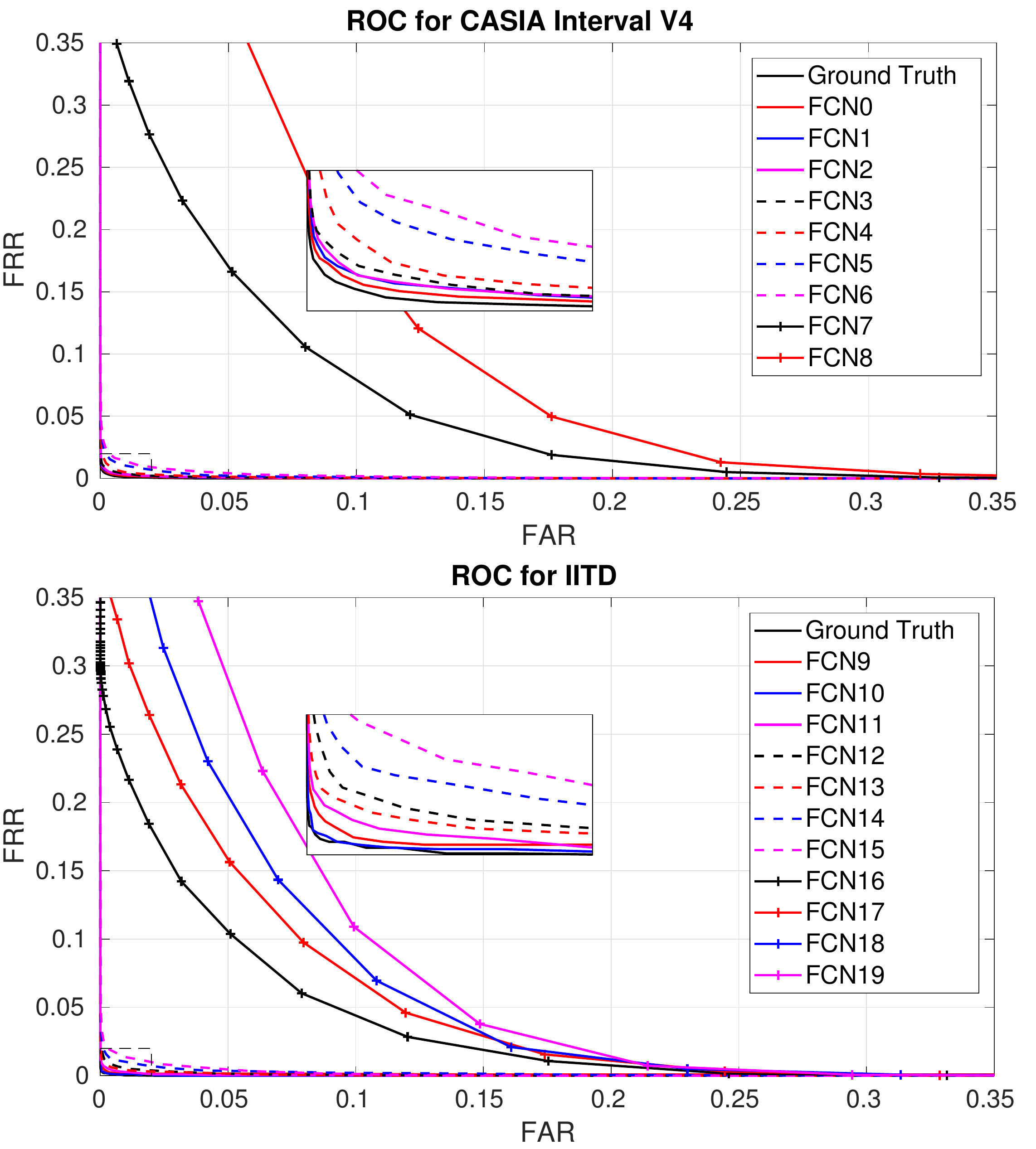}
\end{center}
\vspace{-0.1in}
\caption{Receiver Operating Characteristic (ROC) curves of FCN-based iris recognition pipelines with ground truth segmentation and different FCNs models for CASIA Interval V4 and IITD datasets. The X-axis shows the false acceptance rate, and the Y-axis shows the false rejection rate. In the legend of each dataset, the FCN models are arranged in increasing FLOPs from bottom to top. The zoom-in axis range is [0 0.02] for both x and y directions.}
\label{fig:roc}
\vspace{-0.1in}
\end{figure*}

Figure \ref{fig:roc} shows the ROC curves for the two datasets. Here, the ground truth results are obtained by using the segmentation from IRISSEG-EP \cite{hofbauer2014ground} along with the rest of our flow, which includes contour fitting, normalization, and encoding. As evidenced here, our best performing models achieve ROC close to the ground truth. The EER along with the $\mathcal{F}$-measure results for the models are reported in Table \ref{tbl:eer}. We also provide comparison to previous methods, CAHT \cite{caht} and IrisSeg \cite{irisseg}. We observe that the ground truth EER for each dataset computed using our flow is slightly lower than that reported in IrisSeg. While we cannot provide exact explanation for this result without full access to their experimental setup, we suspect that our contour fitting step might be the reason for the difference since both of the studies use Daugman's normalization and encoding methods.

\begin{table}[!tb]
\caption{Equal Error Rate (EER) and segmentation accuracy ($\mathcal{F}$-measure) comparison between previous approaches, our FCN-based pipeline and groundtruth (GT). In each dataset, FCN models are arranged in increasing FLOPs and $\mathcal{F}$-measure from top to bottom.}
\label{tbl:eer}
\centering
\begin{tabular}{|c|ccc|c|ccc|}
\hline
\multicolumn{4}{|c|}{\textbf{CASIA Interval V4}} & \multicolumn{4}{|c|}{\textbf{IITD}}\\
\hline
\hline
Approach & EER (\%) & $\mathcal{F}$-measure & GFLOPs & Approach & EER (\%) & $\mathcal{F}$-measure & GFLOPs \\ 
\hline
CAHT \cite{caht} & 0.78  & 89.27 & -- & CAHT \cite{caht} & 0.68  & 86.28 & --\\
IrisSeg \cite{irisseg} & 0.62 & 93.10 & -- & IrisSeg \cite{irisseg} & 0.50  & 94.37 & --\\
FCN5 & 0.94 & 98.75 & 0.016 & FCN16 & 6.96 & 97.94 & 0.014\\
FCN4 & 0.64 & 98.93 & 0.060 & FCN15 & 1.13 & 98.15 & 0.038\\
FCN3 & 0.50 & 99.06 & 0.132 & FCN14 & 0.82 & 98.24 & 0.054\\
FCN2 & 0.43 & 99.09 & 0.380 & FCN13 & 0.50 & 98.35 & 0.117\\
FCN1 & 0.42 & 99.14 & 0.513 & FCN12 & 0.60 & 98.38 & 0.154\\
FCN0 & \textbf{0.38} & \textbf{99.17} & 1.143 & FCN11 & 0.41 & 98.50 & 0.335\\
& & & & FCN10 & \textbf{0.19} & 98.59 & 0.453\\
& & & & FCN9 & 0.29 & \textbf{98.62} & 1.791\\
GT & 0.31 & -- & -- & GT & 0.16 & -- & --\\
\hline
\hline
\end{tabular}
\vspace{-0.1in}
\end{table}

\begin{table}[tb]
\caption{Equal Error Rate (EER) and segmentation accuracy ($\mathcal{F}$-measure) comparison between the groundtruth (GT), floating-point, and DFP FCN-based recognition pipelines using the IITD dataset.}
\label{tbl:eer_dfp}
\centering
\begin{tabular}{c|cc|cc|}
\cline{2-5}
& \multicolumn{2}{|c|}{\textbf{Floating-Point}} & \multicolumn{2}{|c|}{\textbf{DFP}}\\
\hline
Model & EER (\%) & $\mathcal{F}$-measure & EER (\%) & $\mathcal{F}$-measure \\ 
\hline
FCN13 & 0.50 & 98.35 & 0.46 & 97.23\\
FCN12 & 0.60 & 98.38 & 0.68 & 96.49\\
FCN11 & 0.41 & 98.50 & 0.22 & 97.24\\
FCN10 & \textbf{0.19} & 98.59 & 0.23 & 96.97\\
FCN9 & 0.29 & 98.62 & 0.37 & 97.14\\
GT & 0.16 & -- & -- & --\\
\hline
\end{tabular}
\vspace{-0.1in}
\end{table}

The results in Table \ref{tbl:eer} show that a few of our FCN models in each dataset outperform previous state-of-the-art EER results from IrisSeg \cite{irisseg}. For CASIA Interval V4, FCN0--FCN3 outperform IrisSeg with FCN0 reducing the EER by almost half. For IITD dataset, FCN9--FCN11 surpass the previous methods with FCN9 reducing EER by more than half. However, it is interesting to note that some of our models achieve significantly higher segmentation accuracy than both CAHT and IrisSeg, while at the same time, these models underperform the previous methods recognition performance. This discrepancy can be attributed to the nature of FCN-based segmentation, which does not strongly account for fine-grained pupil and limbic boundaries labeling. This problem can throw off the contour fitting module in the next stage producing inaccurate center points and radii. This highlights the necessity to evaluate FCN-based design using end-to-end flow rather than segmentation accuracy alone. In future work, this problem may be remedied by assigning larger loss to boundary pixels in comparison to other pixels.

Another evidence for the necessity to perform end-to-end evaluation is between FCN9 and FCN10, where the model with more than 3$\times$ computational complexity and higher segmentation accuracy performs worse in overall recognition performance. This observation is also true for between FCN12 and FCN13. Figure \ref{fig:roc} also verifies this observation where the ROC curves for FCN10 and FCN13 fall below those of FCN9 and FCN12 respectively.

\subsection{Comparisons between DFP and Floating-Point}
Table \ref{tbl:eer_dfp} shows the segmentation accuracy and end-to-end recognition rate comparisons between our floating-point FCN-based pipeline and their DFP counter part. The DFP version of each FCN model is obtained by analyzing and finetuning the trained floating-point weights. From the results in the table, it is evidenced that the quantization process negatively impacts the segmentation accuracy of the models. However, in many cases, the quantization, in fact, improves the overall recognition rates. For instance, for FCN11 and FCN13 the EER improves significantly after the quantization to DFP.

\subsection{Runtime Performance and Hardware Acceleration Speedup}

\begin{figure*}[tb]
\centering
\includegraphics[scale=0.45, trim=0.1cm 0cm 0.1cm 0cm,clip]{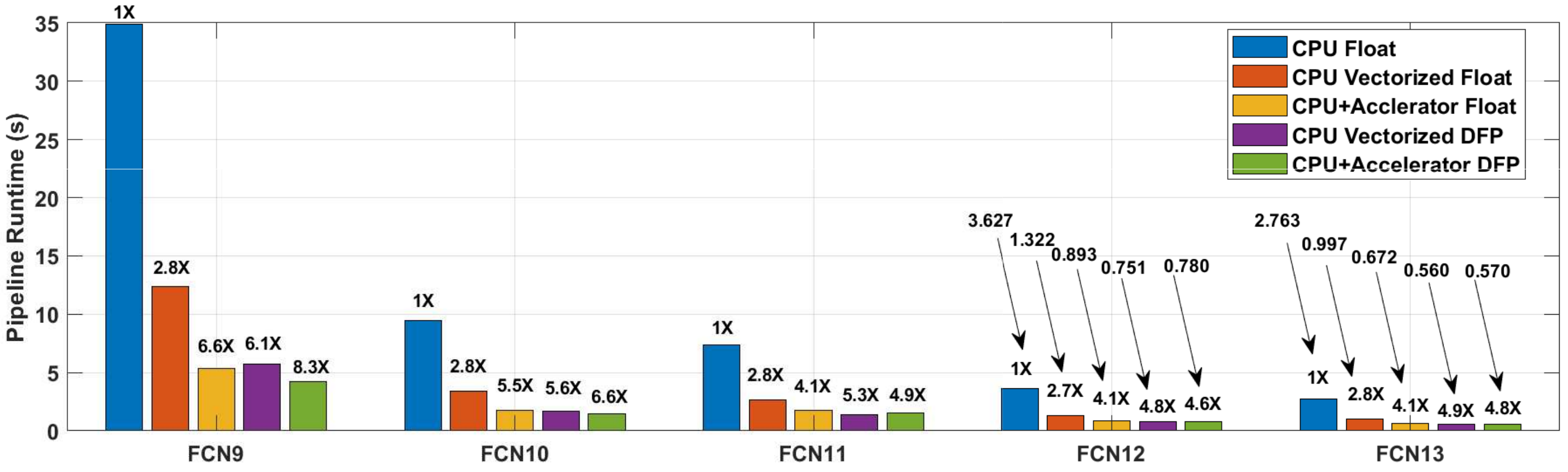}
\caption{Runtime results for end-to-end FCN-based iris recognition pipelines based on different FCN segmentation models for the IITD dataset. Five platform configurations are reported: pure none-vectorized floating-point software (CPU Float), vectorized float-point and fixed-point software using ARM NEON instructions (CPU Vectorized Float, CPU Vectorized DFP) and hardware accelerated with floating-point and DFP acceleerators (CPU+Accelerator Float, CPU+Accelerator DFP). The speedup relative to SW Float is reported on top of each bar.}
\label{fig:speedup}
\vspace{-0.1in}
\end{figure*}

We report the runtime performance of our FCN-based iris recognition pipelines using various FCN models in Figure \ref{fig:speedup}. Due to space constraint, we only report results for FCN9--FCN16 for the IITD dataset. Similar trends and conclusions are observed for FCN0--FCN8 for the CASIA4 Interval dataset. Each runtime result is composed of four components, namely segmentation, contour fitting, normalization and encoding. For each FCN model, we report results for three configurations namely, pure software, vectorized software and hardware accelerated design using our custom accelerator. As discussed in Section \ref{ssec:implementation}, contour fitting, normalization and encoding are always run using pure software. For contour fitting, there are small variations between different input images and FCN models; however, the average runtime is approximately constant across the different runs. Hence, the bulk of the differences among the pipelines stem from the segmentation runtimes  using different FCN models.

\subsubsection{Runtime Results for FPGA SoC} 

In comparison to none-vectorized software, vectorization using the NEON instruction allows between 2.5$\times$ to 2.8$\times$ speedup. Using our accelerator design, we achieve between $2.4\times$ and $6.6\times$ speedup. We observe that higher speedup is realized for larger FCN models since the fraction of runtime spent in segmentation far exceeds that of other components. For the hardware-accelerated implementation, the runtime differences between different FCN pipelines vary by up to two orders of magnitudes, ranging from 0.05s to 5.3s . 

\begin{figure}[tb]
\centering
\subfloat[Floating-Point]{\includegraphics[scale=0.35, trim=4.8cm 1cm 4.7cm 1cm,clip]{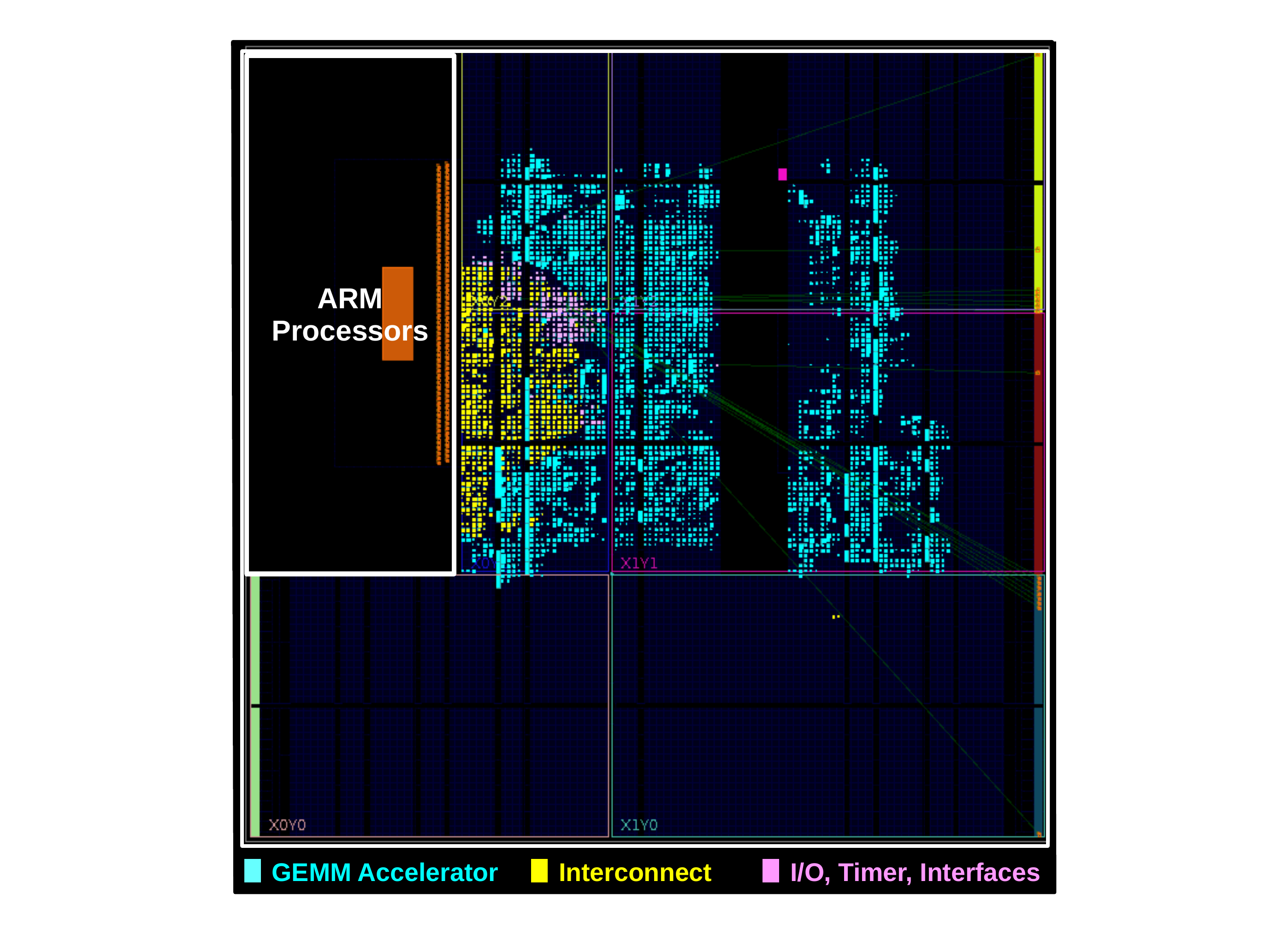}}
\subfloat[Dynamic Fixed-Point]{\includegraphics[scale=0.35, trim=4.8cm 1cm 4.7cm 1cm,clip]{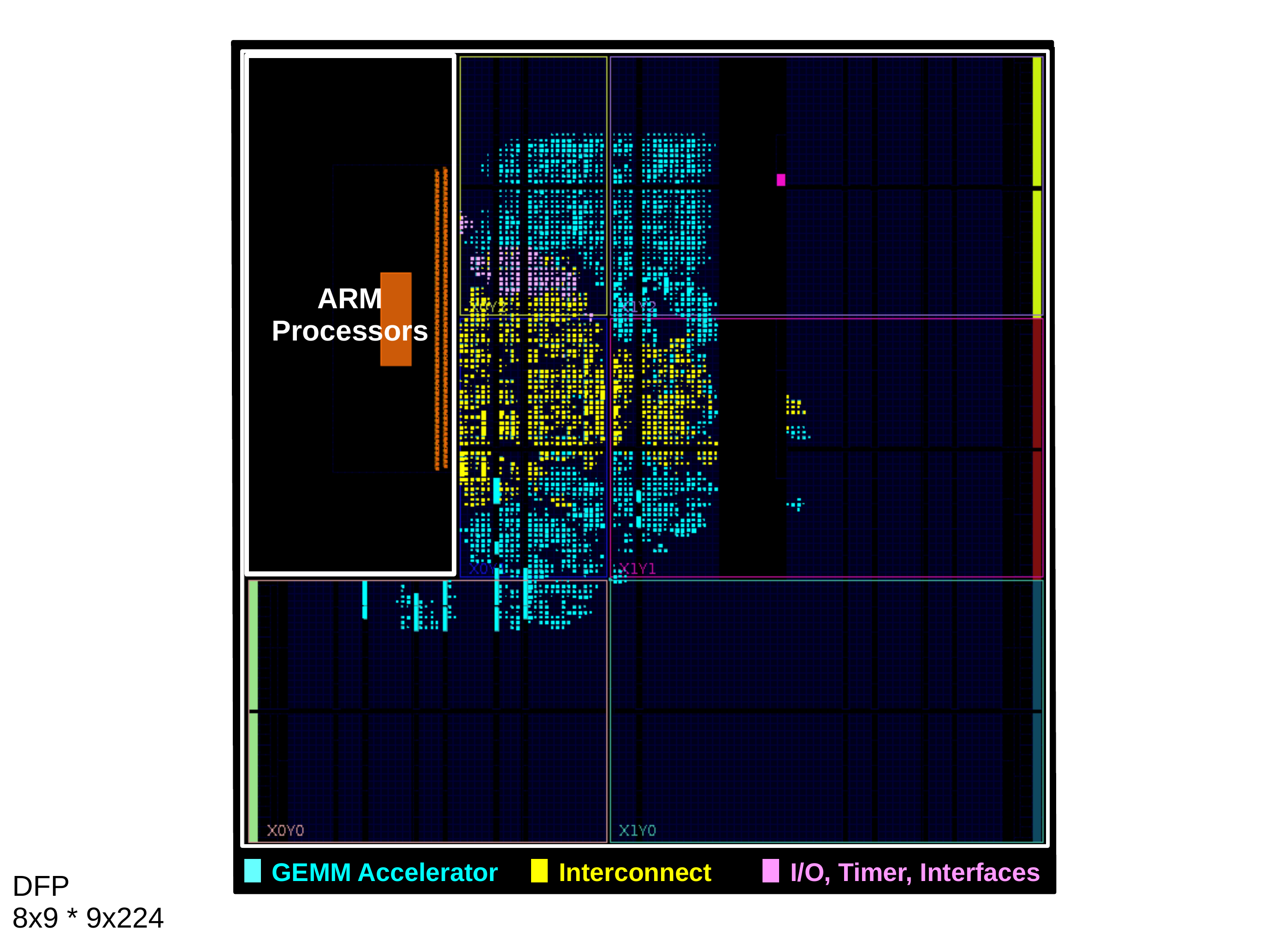}}
\caption{FPGA floorplans of our synthesized accelerators and system modules.}
\label{fig:floorplan}
\end{figure}

The resource utilization of our accelerators is reported in Table \ref{tbl:utilization}, and the floorplans of the designs are shown in Figure \ref{fig:floorplan}. As discussed earlier, since our target models vary significantly in architecture and computational requirement, we implement the accelerators using only the greatest common divisor among them, which explains the low resource utilization. However, with this design, we demonstrate that significant speedup can be achieved while only utilizing a fraction of the available resource. Once a specific model is chosen, a potentially larger speedup can be achieved by optimizing the accelerator design and parameters.

As expected, we observe that overall the floating-point accelerator consumes more resources than the DFP counterpart. Specifically, the floating-point accelerator requires 4$\times$ more DSP resources than fixed-point. While there is a smaller difference in LUT counts, this is due to the required shifting and saturation logic required in the DFP accelerator. For BRAM, the two accelerators utilize the same amount since we require multiple ports for parallel multiplications and accumulations.

\begin{table}[tb]
\caption{FPGA resource utilization for the floating-point and DFP accelerators. The resources include Look-up Tables (LUT), LUT as memory (LUTRAM), Flip-Flop Registers, Block RAM (BRAM), Digital Signal Processing units (DSP), and Global Clock Buffers (BUFG).}
\label{tbl:utilization}
\centering
\begin{tabular}{c|c|c|c|c|c|c|}
\cline{2-7}
& LUT & LUTRAM & Flip-Flop & BRAM & DSP & BUFG\\
\hline
\hline
Floating Point & 15\% & 3\% & 9\% & 5\% & 21\% & 3\%\\
DFP            & 13\% & 2\% & 7\% & 5\% & 5\% & 3\% \\
\hline
\end{tabular}
\vspace{-0.1in}
\end{table}

\subsubsection{Comparison with Embedded GPU} 

For comparison, we also implement our iris recognition pipeline on a Jetson TX1 embedded GPU platform. Table~\ref{tbl:gpu} provides the runtime comparisons for the end-to-end flow between the embedded FPGA and GPU systems. The results show that the GPU perform significantly better than the FPGA platform for larger models such as FCN9 and FCN10. This performance difference can be attributed to the higher operating frequency and more computational resources such as cores and memory bandwidth on the GPU platform. This, however, results in GPU consuming more than double the power requirement for the FPGA platform. In this case, the platform of choice is therefore dependent on the runtime, and energy constraints of the target deployment. For smaller models, surprising runtime results are observed for the GPU platform. From FCN11 to FCN13, the runtime did not decrease as the models become simpler.  Our profiling using Nvidia's {\it nvprof} and {\it Ninsight Systems} shows that most of the runtime is spent in GPU memory allocation and movement. This results in GPU having better energy efficiency for larger models but significantly less efficiency for smaller ones. However, an important note is that the GPU SoC was fabricated with more recent process node of 20nm, which means that for the same 28nm technology node as the FPGA system, the GPU would consume more energy than the results reported in Table~\ref{tbl:gpu}.

\begin{table}[tb!]
\caption{Runtime (in seconds) and energy (in joules) comparison for end-to-end iris recognition flows between embedded FPGA and GPU platforms. The best runtime result for each model is shown in bold. The GPU SoC is based on 20nm process node and the FPGA SoC is based on 28nm.}
\label{tbl:gpu}
\centering
\begin{tabular}{|c|cc|cc|cc|}
\hline
\multirow{2}{*}{Model} & \multicolumn{2}{|c|}{CPU+Accel (Float)} & \multicolumn{2}{|c|}{CPU+Accel (DFP)} &  \multicolumn{2}{|c|}{CPU+GPU (Float)} \\
\cline{2-7}
& Runtime (s) & Energy (J) & Runtime (s) & Energy (J) & Runtime (s) & Energy (J)\\
\hline
FCN13 & 0.67 & 3.35 & {\bf 0.57} & 2.85 & 0.77 & 11.55 \\
FCN12 & 0.89 & 4.45 & {\bf 0.78} & 3.90 & 0.79 & 11.85 \\
FCN11 & 1.79 & 8.95 & 1.51 & 7.55 & {\bf 0.76} & 11.4 \\
FCN10 & 1.73 & 8.65 & 1.43 & 7.15 & {\bf 0.83} & 12.5 \\
FCN9 & 5.32 & 26.6 & 4.20 & 21.0 & {\bf 1.06} & 15.9 \\
\hline
\end{tabular}
\vspace{-0.1in}
\end{table}

\section{Conclusion} \label{sec:conclusion}
In this work, we proposed an end-to-end iris recognition application with FCN-based segmentation. Through our profiling of the overall processing pipeline, we identified that the majority of the runtime is spent on the segmentation step, which was the FCN inference. Targeting this processing stage, we introduced a three-step SW/HW co-design methodology to cut down its runtime. First, we introduced a design space exploration for the FCN architecture to select the most efficient set of models. The exploration was performed through a grid search on several architectural parameters including the spatial dimensions of the input image. For each architecture, we evaluated its segmentation accuracy performance as well as the computational overheads of each FCN model. We then identified the most efficient set of models, which formed a Pareto front. Compared to the FCN architectures from previous works, our best-performing models set new state-of-the-art segmentation accuracy on two well-known datasets, namely CASIA Iris Interval V4 and IITD, while being 50$\times$ more resource efficient. Furthermore, we evaluated the true recognition rate of each model using the end-to-end pipeline and showed that the models outperformed the recognition rate from previous works on the two datasets. Our architectural exploration in this design process showed that a small EER increase of 0.7\% can be traded off for orders of magnitude reduction in computational complexities and latency. With this set of models, we co-designed their datatype to dynamic fixed-point formats for hardware-friendly execution. Finally, we introduced a novel FPGA-based dynamic fixed-point accelerator and demonstrated a full implementation of an accelerated processing flow on an embedded FPGA SoC. We also synthesized a floating-point version of the accelerator for runtime and resources comparisons. In comparison to the onboard CPU, our accelerator is able to achieve up to 8.3$\times$ speedup for the overall pipeline while using only a small fraction of the available FPGA resource. Finally, we provided comparisons between the FPGA system and an embedded GPU showing the different benefits of the two platforms and interesting insights for smaller FCN models. We release a MATLAB-version of our iris recognition flow on Github\footnote{https://github.com/scale-lab/FCNiris}.

The design methodology proposed in this work opens many doors for research on end-to-end systems similar to the demonstrated iris recognition. Future work includes further FCN optimization through pruning, accelerator support for sparse matrix, and improved the contour fitting for none-circular irises using active contour.

\begin{acks}
We would like to thank Dr. Soheil Hashemi for helpful discussions. We also thank NVIDIA Corporation for generous GPU donation. This work is supported by NSF grant 1814920.
\end{acks}

\bibliographystyle{ACM-Reference-Format}
\bibliography{main}

%

\end{document}